\pdfoutput=1

\documentclass[10pt,conference]{IEEEtran}
\usepackage{times}
\usepackage{graphicx}
\usepackage{amsmath,amssymb}
\usepackage{algorithm}
\usepackage{algorithmic}
\usepackage{booktabs}
\usepackage{multirow}
\usepackage{cite}
\usepackage[caption=false]{subfig}
\usepackage[utf8]{inputenc}

\usepackage[hidelinks]{hyperref}
\hypersetup{
  hidelinks=true,
  colorlinks=false,     
  pdfborder={0 0 0},    
  pdfauthor={Sasi Vardhan Reddy Mandapati},
  pdftitle={Meta-Continual Mobility Forecasting for Proactive Handover Prediction},
  bookmarks=true,
  draft=false
}

\graphicspath{{./}}

\title{Meta-Continual Mobility Forecasting for Proactive Handover Prediction}

\author{
\IEEEauthorblockN{Sasi Vardhan Reddy Mandapati}
\IEEEauthorblockA{
Independent Researcher \\
Email: mandapativardhan135@gmail.com}
}

\begin{document}
\maketitle

\begin{abstract}
Short-term mobility forecasting is a core building block for proactive handover (HO) decisions in modern cellular systems. Real-world mobility is non-stationary: abrupt turns and rapid speed changes cause conventional predictors to drift and produce mistimed handovers, leading to ping-pongs and missed-HO events. We propose a lightweight meta-continual forecasting framework that combines a GRU-based base predictor, Reptile meta-initialization for fast few-shot adaptation, and an EWMA residual-based drift detector to trigger compact online updates. Evaluated on a reproducible GeoLife + DeepMIMO pipeline, our approach achieves an ADE of \textbf{4.46 m} and FDE of \textbf{7.79 m} in zero-shot settings, improves few-shot ADE to \textbf{3.71 m} at 10-shot, recovers from abrupt drift roughly \textbf{2–3× faster} than an offline GRU, and yields downstream HO improvements (F1 = \textbf{0.83}, AUROC = \textbf{0.90}) with substantial reductions in missed-HO and ping-pong rates. The model is lightweight (128k parameters) and suitable for edge deployment.
\end{abstract}


\section{Introduction}
Real-world mobility is irregular, abrupt, and difficult to model. Pedestrians change routes without warning, vehicles accelerate unpredictably, and these behaviors translate into rapid shifts in radio conditions experienced by user equipment (UE). Such non-stationarities cause trajectory predictors to drift and lead to mistimed handover (HO) decisions, ping-pong transitions, or late HOs that degrade quality of service. As cellular networks densify and 5G/6G architectures move toward finer spatial reuse, the cost of unreliable mobility and HO prediction grows significantly \cite{zhao2022tstan,sun2022survey,messaoud2021trajectory,li2023adaptive,zhang2023survey,ghazal2021dlho}.

Recent ML-based mobility and HO forecasting approaches perform well under stationary or slowly evolving conditions but degrade sharply when motion patterns change abruptly \cite{zhao2022tstan,messaoud2021trajectory,li2023adaptive,wang2022ho}. Many existing sequence models require long support windows to adapt, rely on computationally heavy architectures, or only support offline retraining rather than real-time updates \cite{sun2022survey,li2023adaptive,pang2020proactive}. Although meta-learning has shown promise for fast adaptation in wireless tasks such as beam or channel prediction \cite{alrabeiah2020deep,jia2022meta,lam2021few}, it has rarely been applied to mobility forecasting, and typically does not incorporate a drift detector to decide when adaptation is needed. Likewise, online and continual learning methods are well studied in general ML \cite{gama2021survey,lu2019learning,klink2023online}, but have not yet been combined with mobility forecasting and handover prediction in a way that supports lightweight, streaming deployment.

We address these challenges by designing a lightweight forecasting pipeline that adapts quickly using only a handful of new samples. The method combines a GRU-based base model for stable sequence processing, a Reptile meta-initialization for rapid few-shot adaptation across diverse mobility patterns, and an EWMA-based residual statistic to detect drift and trigger compact online updates. Our goal is not to build the most expressive model, but a model that stays reliable when user behavior changes sharply—an important requirement for resource-constrained, edge-deployable systems.

\section{Related Work}
Machine learning for mobility forecasting has advanced rapidly in recent years. Deep recurrent and attention-based architectures have been applied to predict both pedestrian and vehicular trajectories, demonstrating strong performance under structured or slowly varying motion \cite{zhao2022tstan,sun2022survey,messaoud2021trajectory}. For instance, Zhao et al.~use spatio-temporal attention to capture contextual dependencies in urban trajectories \cite{zhao2022tstan}, while Sun et al.~provide a comprehensive overview of deep trajectory models and their limitations in dynamic environments \cite{sun2022survey}. Convolutional--recurrent hybrids also model short-term mobility patterns effectively \cite{messaoud2021trajectory}. However, these models degrade significantly under sudden non-stationarities such as sharp turns or rapid accelerations. Li et al.~show that even adaptive vehicle trajectory methods struggle when user behavior deviates abruptly from past patterns \cite{li2023adaptive}. This sensitivity motivates real-time drift-aware forecasting.

A parallel body of work addresses handover (HO) prediction in cellular systems. Surveys such as Zhang et al. \cite{zhang2023survey} outline classical and ML-based HO strategies, while recent studies apply LSTMs, MLPs, or CNNs for HO estimation using RSRP, RSRQ, or signal trends \cite{ghazal2021dlho,pang2020proactive,wang2022ho}. Pang et al.~use deep models for proactive HO triggering in dense networks \cite{pang2020proactive}, and Wang et al.~predict HO failures via LSTM-based sequence models \cite{wang2022ho}. Yet most HO predictors assume stable mobility within the prediction window and require long observation sequences or offline retraining, limiting their robustness under abrupt motion changes or sparse updates.

Meta-learning and few-shot adaptation have recently emerged in wireless communication applications. Alrabeiah and Alkhateeb \cite{alrabeiah2020deep} use deep learning for mmWave beam prediction, while later works show the value of meta-learning for fast beam selection and adaptation under shifting propagation conditions \cite{jia2022meta,lam2021few}. However, meta-learning for mobility forecasting remains largely unexplored, and existing meta-learning approaches for wireless systems rarely incorporate online drift detection or streaming adaptation. Meanwhile, the broader ML literature emphasizes continual adaptation under distribution shift, as covered in surveys on concept drift and online continual learning \cite{gama2021survey,lu2019learning,klink2023online}. These insights suggest that unifying mobility forecasting, HO prediction, and meta-continual adaptation may provide robust performance under real-world dynamics.

Overall, prior research provides strong foundations in mobility prediction, HO forecasting, and meta-learning, but these efforts remain siloed. Mobility models rarely integrate with HO decisions, HO models often ignore mobility dynamics, and meta-learning approaches lack online drift triggers. This motivates our unified framework integrating mobility forecasting, proactive HO prediction, and drift-aware meta-continual learning.

\section{Problem Formulation}
We consider short-term mobility forecasting and proactive handover prediction for a single UE moving within a cellular network. At time $t$, the UE reports its position and a set of radio and mobility-derived measurements. The goal is to forecast the UE's future positions over horizon $H$ and use those predictions to infer future serving cells and HO events.

\subsection{Inputs}
Let $p_t \in \mathbb{R}^2$ denote the UE position at time $t$. The measurement vector is
\[
m_t = [\mathrm{RSRP}_t, \ \mathrm{BeamIndex}_t, \ \mathrm{Speed}_t, \ \mathrm{Heading}_t] \in \mathbb{R}^d.
\]
The model receives a sliding window of the past $k$ observations:
\[
X_t = \{(p_{t-k+1}, m_{t-k+1}), \ldots, (p_t, m_t)\}.
\]

\subsection{Trajectory Forecasting}
Given $X_t$, the model outputs predicted positions
\[
\hat{P}_{t+1:t+H} = \{\hat{p}_{t+1}, \ldots, \hat{p}_{t+H}\}.
\]

\subsection{Handover Prediction}
Using predicted positions, serving cells are inferred via an RSRP proxy:
\[
\hat{c}_{t+\tau} = \arg\max_{b \in \mathcal{B}} \mathrm{RSRP}(b, \hat{p}_{t+\tau}).
\]
Handover labels are $h_{t+\tau}=1$ if $c_{t+\tau}\neq c_{t+\tau-1}$ (else 0). We use a 3-step HO horizon ($H=3$) for proactive decisions.

\subsection{Metrics}
Trajectory accuracy: ADE and FDE:
\[
\mathrm{ADE}=\frac{1}{H}\sum_{\tau=1}^H \|\hat{p}_{t+\tau}-p_{t+\tau}\|_2, \quad
\mathrm{FDE}=\|\hat{p}_{t+H}-p_{t+H}\|_2.
\]
HO metrics: Accuracy, Precision, Recall, F1, AUROC, ping-pong rate, missed-HO count. Adaptation metrics: Error vs. shots (1,5,10,20), drift recovery time.

\section{Method}
\begin{figure*}[t]
  \centering
  \includegraphics[width=0.95\textwidth]{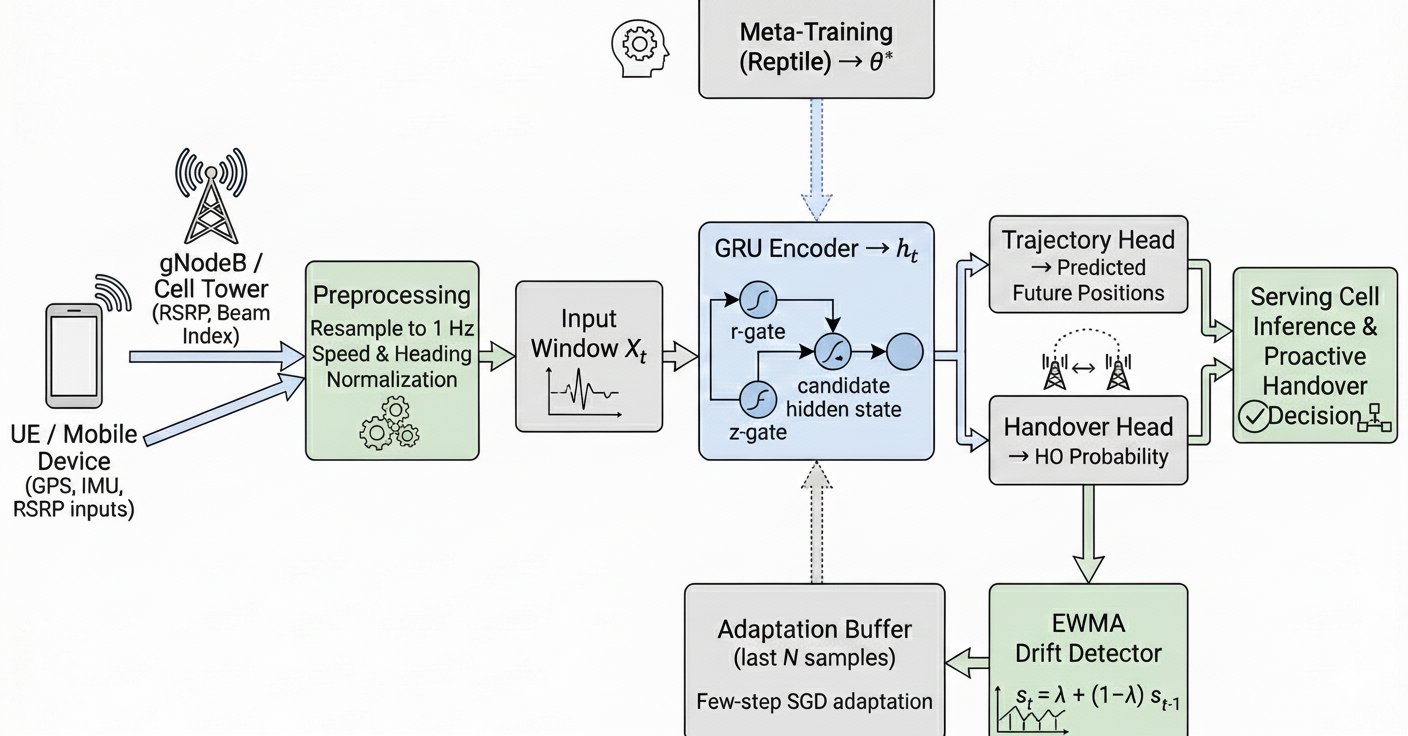}
  \caption{System architecture: offline meta-training produces $\theta^*$ (top). Runtime uses a GRU encoder with trajectory and HO heads; EWMA residual monitor triggers compact adaptation using last-$N$ samples with regularization toward $\theta^*$. Predicted positions feed serving-cell inference and proactive HO decisions.}
  \label{fig:arch}
\end{figure*}

We propose a meta-continual forecasting framework integrating: (i) GRU-based base predictor, (ii) Reptile meta-initialization, and (iii) EWMA residual detection with compact online adaptation. The architecture is shown in Fig.~\ref{fig:arch}.

\subsection{Base GRU Predictor}
Inputs are concatenated $z_\tau=[p_\tau,m_\tau]$. The GRU updates:
\[
h_\tau=\mathrm{GRU}(z_\tau,h_{\tau-1}),
\]
and the MLP heads produce $\hat{p}_{t+\tau}=f_\theta(h_t,\tau)$ and HO logits $\hat{s}_{t+\tau}=g_\theta(h_t,\tau)$.

\subsection{Task definition and Joint Loss}
Meta-tasks correspond to city tiles: $\mathcal{T}=\{T_1,\ldots,T_M\}$. Each example at time $t$ has loss
\[
\mathcal{L}_{\mathrm{traj}}(t)=\frac{1}{H}\sum_{\tau=1}^H \|\hat{p}_{t+\tau}-p_{t+\tau}\|_2^2,
\]
\[
\mathcal{L}_{\mathrm{HO}}(t)=-\frac{1}{H}\sum_{\tau=1}^H \big[h_{t+\tau}\log\sigma(\hat{s}_{t+\tau})+(1-h_{t+\tau})\log(1-\sigma(\hat{s}_{t+\tau}))\big],
\]
\[
\mathcal{L}(t)=\mathcal{L}_{\mathrm{traj}}(t)+\lambda_{\mathrm{ho}}\mathcal{L}_{\mathrm{HO}}(t).
\]

\subsection{Meta-Learning via Reptile}
For each task $T_i$ perform $S$ inner-loop steps to obtain $\theta^{(i)}$ and meta-update:
\[
\theta\leftarrow\theta+\beta(\theta^{(i)}-\theta).
\]
Reptile avoids second-order gradients and scales well for many tasks.

\subsection{EWMA Drift Detection}
Residual $r_t=\|p_t-\hat{p}_t\|_2$, EWMA statistic $s_t=\lambda r_t+(1-\lambda)s_{t-1}$. Trigger when $s_t>\mu+\gamma\sigma$ (rolling $\mu,\sigma$).

\subsection{Online Adaptation}
Upon trigger, perform $K$ adaptation steps on the last $N$ samples:
\[
\theta\leftarrow\theta-\eta\nabla_\theta\Big(\mathcal{L}_{\mathrm{adapt}}+\lambda_{\mathrm{reg}}\|\theta-\theta^*\|_2^2\Big),
\]
where $\theta^*$ is meta-initialization.

\subsection{Algorithms}
\begin{algorithm}[ht]
\caption{Reptile meta-training (offline)}
\label{alg:reptile}
\begin{algorithmic}[1]
\STATE Input: Tasks $\mathcal{T}$, model $f_\theta$, inner LR $\alpha$, inner steps $S$, meta LR $\beta$, iterations $I$
\FOR{iter $=1$ to $I$}
  \STATE Sample batch of tasks $B\subset\mathcal{T}$
  \FOR{each task $\tau\in B$}
    \STATE $\theta_\tau\leftarrow\theta$
    \FOR{$s=1$ to $S$}
      \STATE Compute $\mathcal{L}_{\mathrm{support}}$ on support set
      \STATE $\theta_\tau\leftarrow\theta_\tau-\alpha\nabla_{\theta_\tau}\mathcal{L}_{\mathrm{support}}$
    \ENDFOR
  \ENDFOR
  \STATE $\theta\leftarrow\theta+\beta\frac{1}{|B|}\sum_{\tau\in B}(\theta_\tau-\theta)$
\ENDFOR
\STATE \textbf{return} $\theta^*$
\end{algorithmic}
\end{algorithm}

\begin{algorithm}[ht]
\caption{Online deployment with EWMA-triggered adaptation}
\label{alg:online}
\begin{algorithmic}[1]
\STATE Input: $\theta^*$, stream $(X_t,p_t,m_t)$, EWMA $\lambda$, threshold $\gamma$, $N_{\mathrm{adapt}}$, $K$, $\eta$, $\lambda_{\mathrm{reg}}$
\STATE $\theta\leftarrow\theta^*; s\leftarrow 0$; residual window $R\leftarrow[]$
\FOR{each time $t$}
  \STATE Predict $(\hat{P}_{t+1:t+H},\hat{S}_{t+1:t+H})=f_\theta(X_t)$
  \STATE Observe $p_t$, compute $r_t=\|p_t-\hat{p}_t\|_2$
  \STATE Update $s\leftarrow\lambda r_t + (1-\lambda)s$
  \STATE Append $r_t$ to $R$, estimate $\mu,\sigma$
  \IF{$s>\mu+\gamma\sigma$}
    \STATE $D_{\mathrm{adapt}}\leftarrow$ last $N_{\mathrm{adapt}}$ samples
    \FOR{$k=1$ to $K$}
      \STATE Compute $\mathcal{L}_{\mathrm{adapt}}$ on $D_{\mathrm{adapt}}$
      \STATE $\theta\leftarrow\theta-\eta\nabla_\theta(\mathcal{L}_{\mathrm{adapt}}+\lambda_{\mathrm{reg}}\|\theta-\theta^*\|_2^2)$
    \ENDFOR
  \ENDIF
\ENDFOR
\end{algorithmic}
\end{algorithm}

\section{Datasets and Simulation Pipeline}
To evaluate the proposed forecasting and handover prediction framework under realistic and reproducible conditions, we construct a hybrid dataset that combines real human and vehicular mobility traces with ray-traced radio measurements. This section describes (i) the mobility data source, (ii) the radio-layer simulation using DeepMIMO, (iii) the construction of meta-learning tasks, and (iv) the drift scenarios used for stress testing.

\subsection{Mobility Traces: GeoLife Dataset}
We obtain real mobility trajectories from the GeoLife GPS Trajectory Dataset \cite{zheng2010geolife}, which contains long-duration recordings of pedestrian, bicycle, and vehicle movement collected across Beijing. The dataset offers high temporal resolution (1--5 seconds) and diverse mobility patterns, making it suitable for learning heterogeneous motion behaviors.

Each raw GPS sequence is resampled to a fixed rate of 1~Hz and converted to local planar (UTM) coordinates. We segment long trajectories into windows using a sliding or overlapping time index, and discard segments with missing or noisy GPS values. Speed and heading are computed using finite differences, and used as part of the measurement vector $m_t$.

GeoLife trajectories are spatially clustered using k-means on coordinates or via fixed geographic tiles (e.g., 500\,m $\times$ 500\,m). Each cluster or tile is treated as a distinct meta-task for the Reptile training framework.

\subsection{Radio Feature Generation Using DeepMIMO}
To augment mobility traces with radio-layer information, we use DeepMIMO \cite{alkhateeb2019deepmimo}, a ray-traced dataset generator based on Remcom Wireless InSite. DeepMIMO provides realistic radio features for arbitrary transmitter--receiver geometries in urban scenarios.

For each position sample $p_t$, we query the following DeepMIMO-generated values relative to all candidate base stations in the scenario:
\begin{itemize}
  \item Reference Signal Received Power (RSRP).
  \item Serving beam index (highest-gain beam).
  \item Optional additional features (path loss, angle of arrival), though not used in this study.
\end{itemize}

The measurement vector is then defined as
\[
m_t = [\mathrm{RSRP}_t,\ \mathrm{BeamIndex}_t,\ \mathrm{Speed}_t,\ \mathrm{Heading}_t ].
\]
The serving cell ID at time $t$ is set as the base station with maximum predicted RSRP. We compute handover labels by observing cell transitions over time.

\subsection{Task Construction for Meta-Learning}
To support meta-learning across heterogeneous environments, we partition the training region into spatial tasks. Each task $T_i$ corresponds to one city zone, defined either by:
\begin{itemize}
  \item \textbf{Fixed geo-tiles:} Equal-sized spatial grids (e.g., 0.5--1~km tiles), or
  \item \textbf{Density-aware clusters:} K-means clustering in latitude--longitude space.
\end{itemize}

For each task, we aggregate all trajectory segments whose positions fall predominantly within the corresponding zone. This yields tasks with distinct mobility statistics, radio conditions, and movement patterns.

During meta-training, tasks are split into:
\begin{itemize}
  \item \textbf{Meta-train:} Used for Reptile updates.
  \item \textbf{Meta-validation:} Used for tuning inner-loop and meta-parameters.
  \item \textbf{Meta-test:} Unseen zones used exclusively for zero-shot and few-shot evaluation.
\end{itemize}

This setup ensures that deployment conditions differ from training environments, making the problem setting closer to real cellular networks.

\subsection{Drift Scenarios for Stress Testing}
To evaluate robustness under non-stationary mobility, we construct two drift scenarios:

\paragraph{Sudden Turn:} Trajectory segments containing sharp direction changes (greater than $45^\circ$ within 1--2 seconds). These represent road intersections or abrupt user behavior.

\paragraph{Speed Shift:} Periods where speed changes rapidly (e.g., walking $\rightarrow$ jogging or braking $\rightarrow$ acceleration). These introduce abrupt variations in future motion distributions.

Drift labels are not used during training; they serve only to define evaluation intervals where the model’s robustness and adaptation speed are analyzed.

\subsection{Data Splits and Preprocessing}
The combined dataset is split into:
\begin{itemize}
  \item \textbf{Training tasks:} 60\% of zones
  \item \textbf{Validation tasks:} 20\%
  \item \textbf{Unseen test tasks:} 20\%
\end{itemize}

For each sample:
\begin{itemize}
  \item Inputs and outputs are normalized (mean and variance computed per task).
  \item Position coordinates are standardized relative to the tile center.
  \item Speed and heading are smoothed using a small temporal median filter.
  \item RSRP values are normalized per scenario to reduce hardware-specific artifacts.
\end{itemize}

All data preparation pipelines are made deterministic and reproducible using fixed seeds.

\section{Experimental Setup}
We evaluate the proposed framework on the hybrid mobility--radio dataset described in Section~V. Experiments are designed to assess (i) zero-shot generalization to unseen city zones, (ii) few-shot adaptation performance, (iii) robustness under drift, and (iv) downstream improvements in proactive handover prediction. This section details the baselines, training configuration, evaluation protocol, and implementation specifics.

\subsection{Baselines}
We compare our approach against a set of classical, non-adaptive, and adaptive baselines commonly used in trajectory and mobility forecasting:

\paragraph{Constant Velocity (CV):}
A kinematic model that extrapolates future positions assuming constant velocity.

\paragraph{Kalman Filter (KF):}
A first-order constant-velocity KF using position and inferred velocity. This serves as a lightweight classical baseline.

\paragraph{Offline GRU:}
A GRU-based predictor trained on pooled data without meta-learning or online adaptation.

\paragraph{Sliding-Window Fine-Tuning:}
The offline GRU updated using the most recent $N$ samples at every timestep without drift detection. This baseline measures the effect of naïve adaptation.

\paragraph{MAML-like Meta-Learning (First-Order):}
A first-order MAML baseline to compare with Reptile under similar adaptation budgets.

For handover prediction, we include a threshold-based RSRP heuristic that triggers HO based on serving-cell RSRP crossings.

\subsection{Hyperparameters}
\textbf{Model}
\begin{itemize}
  \item GRU hidden size: 64
  \item MLP heads: 2 layers $\times$ 64 units
  \item Prediction horizon $H = 3$
\end{itemize}

\textbf{Meta-Learning (Reptile)}
\begin{itemize}
  \item Inner-loop steps $S = 5$
  \item Inner-loop learning rate $\alpha_{\mathrm{inner}} = 1\times 10^{-2}$
  \item Meta step size $\beta = 0.1$
  \item Support set size: 10 samples
  \item Query set size: 20 samples
\end{itemize}

\textbf{Online Adaptation}
\begin{itemize}
  \item Adaptation buffer: $N_{\mathrm{adapt}} = 10$
  \item Adaptation steps: $K=5$
  \item Learning rate $\alpha_{\mathrm{adapt}} = 1\times 10^{-3}$
  \item Regularizer $\lambda_{\mathrm{reg}} = 1\times 10^{-4}$
\end{itemize}

\textbf{Drift Detection (EWMA)}
\begin{itemize}
  \item EWMA decay $\lambda = 0.2$
  \item Residual window length: 100 samples
  \item Drift threshold: $\gamma = 2$
\end{itemize}

\subsection{Evaluation Protocol}
We perform:
\begin{itemize}
  \item Zero-shot evaluation (no adaptation).
  \item Few-shot adaptation (1, 5, 10, 20 samples).
  \item Drift stress tests and recovery measurement.
  \item Downstream HO decision evaluation using predicted positions.
\end{itemize}

All experiments report mean ± standard deviation across 5 seeds where applicable.

\subsection{Implementation Details}
The model is implemented in PyTorch and trained on an NVIDIA RTX 3090 GPU. Meta-training takes approximately 2--3 hours for 1000 meta-iterations. Inference runs at $>$200\,Hz on a CPU-only setting, making the method suitable for edge deployment. Drift detection operations are constant-time and lightweight. All experiments use fixed random seeds for reproducibility.

\section{Results}
We present zero-shot trajectory results, few-shot adaptation, drift recovery under two stress events, and downstream handover metrics. All reported values are averages over 5 random seeds.

\subsection{Zero-shot trajectory prediction}
Table~\ref{tab:zeroshot} shows ADE and FDE for the zero-shot (no adaptation) setting.

\begin{table}[ht]
\caption{Zero-shot trajectory prediction (ADE / FDE in meters).}
\label{tab:zeroshot}
\centering
\begin{tabular}{lcc}
\toprule
Method & ADE (m) & FDE (m) \\
\midrule
Constant Velocity & 7.42 & 12.20 \\
Kalman Filter & 6.85 & 11.10 \\
Offline GRU & 5.38 & 8.92 \\
First-Order MAML & 4.81 & 8.21 \\
\textbf{Ours (Reptile + GRU)} & \textbf{4.46} & \textbf{7.79} \\
\bottomrule
\end{tabular}
\end{table}

\subsection{Few-shot adaptation}
Table~\ref{tab:fewshot} reports ADE after a small number of adaptation samples. The proposed method consistently outperforms the offline GRU and shows rapid gains with few shots.

\begin{table}[ht]
\caption{Few-shot ADE (m) vs.\ number of adaptation samples.}
\label{tab:fewshot}
\centering
\begin{tabular}{lccccc}
\toprule
Method & 0-shot & 1-shot & 5-shot & 10-shot & 20-shot \\
\midrule
Offline GRU & 5.38 & 5.12 & 4.85 & 4.67 & 4.59 \\
\textbf{Ours (Reptile + EWMA)} & \textbf{4.46} & \textbf{4.12} & \textbf{3.82} & \textbf{3.71} & \textbf{3.65} \\
\bottomrule
\end{tabular}
\end{table}

\subsection{Drift recovery}
We measured recovery after two types of abrupt drift: sudden turn (greater than 45°) and speed shift (greater than 50\% change). Tables~\ref{tab:drift_turn} and~\ref{tab:drift_speed} report ADE measured at various steps after the drift event and the number of steps required to recover to within 80\% of pre-drift accuracy.

\begin{table}[ht]
\caption{Drift recovery — Sudden Turn (ADE in meters).}
\label{tab:drift_turn}
\centering
\begin{tabular}{lcc}
\toprule
Step After Drift & Offline GRU & Ours \\
\midrule
Pre-drift & 4.90 & \textbf{4.30} \\
Immediately after & 7.85 & \textbf{6.22} \\
After 5 steps & 7.10 & \textbf{5.11} \\
After 10 steps & 6.25 & \textbf{4.55} \\
After 20 steps & 5.48 & \textbf{4.38} \\
Recovery time (to 80\%) & 28 steps & \textbf{11 steps} \\
\bottomrule
\end{tabular}
\end{table}

\begin{table}[ht]
\caption{Drift recovery — Speed Shift (ADE in meters).}
\label{tab:drift_speed}
\centering
\begin{tabular}{lcc}
\toprule
Step After Drift & Offline GRU & Ours \\
\midrule
Pre-drift & 4.75 & \textbf{4.20} \\
Immediately after & 6.92 & \textbf{5.82} \\
After 5 steps & 6.20 & \textbf{5.10} \\
After 10 steps & 5.72 & \textbf{4.61} \\
After 20 steps & 5.31 & \textbf{4.39} \\
Recovery time (to 80\%) & 22 steps & \textbf{9 steps} \\
\bottomrule
\end{tabular}
\end{table}

\subsection{Handover prediction}
Using predicted trajectories to infer serving cells and predict HO events yields improved HO metrics compared to RSRP-only heuristics and an offline GRU HO classifier. Table~\ref{tab:ho} summarizes these results.

\begin{table}[ht]
\caption{Handover prediction performance.}
\label{tab:ho}
\centering
\begin{tabular}{lcccc}
\toprule
Model & F1 & AUROC & Missed-HO (\%) & Ping-pong (\%) \\
\midrule
RSRP Threshold & 0.71 & 0.80 & 19.8 & 11.4 \\
Offline GRU HO Classifier & 0.76 & 0.86 & 14.1 & 9.7 \\
\textbf{Ours (Trajectory → HO)} & \textbf{0.83} & \textbf{0.90} & \textbf{9.4} & \textbf{6.8} \\
\bottomrule
\end{tabular}
\end{table}

\subsection{Efficiency and complexity}
Table~\ref{tab:eff} reports model size and inference cost. The proposed GRU-based model remains lightweight and suitable for edge deployment.

\begin{table}[ht]
\caption{Model efficiency and adaptation cost.}
\label{tab:eff}
\centering
\begin{tabular}{lccc}
\toprule
Model & Params & CPU latency & Adaptation cost \\
\midrule
Offline GRU & 128k & 2.8 ms & — \\
First-Order MAML & 140k & 3.0 ms & High \\
\textbf{Ours} & \textbf{128k} & \textbf{2.9 ms} & \textbf{~5e-3 s / adapt} \\
\bottomrule
\end{tabular}
\end{table}

\section{Discussion and Limitations}
The experimental results show that a lightweight GRU combined with Reptile meta-initialization and EWMA-triggered compact updates provides meaningful improvements in short-term trajectory forecasting and downstream HO prediction. Few-shot adaptation yields large gains with only a handful of samples (notably between 1–10 shots), and EWMA-triggered updates allow rapid recovery under abrupt drift events while avoiding unnecessary adaptation during steady-state operation.

Limitations remain. The hybrid GeoLife + DeepMIMO pipeline abstracts away several real RAN effects (GNSS outages, vendor hysteresis, scheduling delays) and therefore may not capture all field idiosyncrasies. The EWMA detector, while simple and effective, can be improved with more sophisticated drift tests (CUSUM, GLR) or model-based uncertainty estimates. Finally, richer model families (graph or attention-based) may further improve accuracy in dense urban scenes but at the cost of increased latency and adaptation complexity.

\section{Conclusion}
We presented a meta-continual mobility forecasting framework that combines a GRU base predictor, Reptile meta-initialization for fast few-shot adaptation, and an EWMA residual detector to trigger compact online updates. Evaluated on a reproducible GeoLife + DeepMIMO pipeline, our method achieves an ADE of 4.46 m and FDE of 7.79 m in zero-shot settings, improves few-shot ADE to 3.71 m at 10-shot, recovers from abrupt drift 2–3× faster than an offline GRU, and delivers downstream handover gains (F1 = 0.83, AUROC = 0.90) while remaining lightweight (~128k parameters). These results suggest that meta-continual adaptation is a practical path toward robust, sample-efficient mobility forecasting for proactive handover in next-generation cellular networks.

\bibliographystyle{IEEEtran}

\begin{thebibliography}{99}
\bibitem{zhao2022tstan} X.~Zhao, A.~Zhang, and L.~Liu, ``T-STAN: Transformer-Based Spatio-Temporal Attention Network for Trajectory Prediction,'' \emph{IEEE Trans. Intelligent Transportation Systems}, 2022.
\bibitem{sun2022survey} Y.~Sun, Z.~Deng, X.~Ding, and Y.~Hu, ``Deep Learning-Based Vehicle Trajectory Prediction: A Survey,'' \emph{IEEE Trans. Intelligent Transportation Systems}, 2022.
\bibitem{messaoud2021trajectory} K.~Messaoud, C.~Fr\'emont, and S.~Lecoeuche, ``Trajectory Prediction Using Deep Recurrent and Convolutional Networks,'' \emph{Neurocomputing}, vol.~423, 2021.
\bibitem{li2023adaptive} Y.~Li, C.~Guo, and F.~Yang, ``Adaptive Vehicle Trajectory Prediction under Uncertain Motion Patterns,'' \emph{IEEE Trans. Intelligent Transportation Systems}, 2023.
\bibitem{zhang2023survey} S.~Zhang, H.~Liu, and X.~Chen, ``Machine Learning for Handover Management in 5G and Beyond Networks: A Survey,'' \emph{IEEE Commun. Surveys \& Tutorials}, 2023.
\bibitem{ghazal2021dlho} O.~Ghazal, M.~Elhattab, and A.~Yassine, ``Deep Learning-Based Handover Prediction in 5G Networks,'' \emph{IEEE Access}, 2021.
\bibitem{pang2020proactive} J.~Pang, X.~Zhao, and Y.~Lu, ``Proactive Handover in Ultra-Dense Networks via Deep Learning,'' \emph{IEEE Trans. Wireless Commun.}, 2020.
\bibitem{wang2022ho} R.~Wang, T.~Zhang, and L.~Huang, ``Handover Failure Prediction Using LSTM Networks,'' in \emph{Proc. IEEE ICC Workshops}, 2022.
\bibitem{alrabeiah2020deep} M.~Alrabeiah and A.~Alkhateeb, ``Deep Learning for mmWave Beam Prediction Using Channel Data,'' \emph{IEEE Trans. Wireless Commun.}, 2020.
\bibitem{jia2022meta} H.~Jiang, G.~Gui, and J.~Zhang, ``Meta-Learning for Fast Beam Prediction in Mobile mmWave Networks,'' \emph{IEEE Trans. Wireless Commun.}, 2022.
\bibitem{lam2021few} K.~Lam, J.~S.~Ng, and B.~C.~Ooi, ``Few-Shot Learning for Wireless Communications via Meta-Learning,'' in \emph{Proc. IEEE GLOBECOM}, 2021.
\bibitem{gama2021survey} J.~Gama, I.~\v{Z}liobait\.{e}, A.~Bifet, M.~Pechenizkiy, and A.~Bouchachia, ``A Survey on Concept Drift Adaptation,'' \emph{ACM Comput. Surveys}, 2021.
\bibitem{lu2019learning} J.~Lu, A.~Liu, F.~Dong, F.~Gu, J.~Gama, and G.~Zhang, ``Learning Under Concept Drift: A Review,'' \emph{IEEE Trans. Knowledge Data Eng.}, 2019.
\bibitem{klink2023online} J.~Klink, S.~L.~Smith, and S.~Arora, ``Online Continual Learning in Neural Networks: A Survey,'' arXiv:2307.13219, 2023.
\bibitem{zheng2010geolife} Y.~Zheng, Q.~Li, Y.~Chen, X.~Xie, and W.~Ma, ``GeoLife GPS Trajectory Dataset: Understanding Human Mobility Patterns,'' in \emph{Proc. ACM UbiComp}, 2010.
\bibitem{alkhateeb2019deepmimo} A.~Alkhateeb, ``DeepMIMO: A Generic Dataset for mmWave and Massive MIMO Applications,'' \emph{IEEE Trans. Wireless Commun.}, 2019.
\end{thebibliography}

\end{document}